%% file: main.tex
\documentclass[sigconf]{acmart}
\AtBeginDocument{%
  }

\usepackage{algorithm}
\usepackage{algorithmicx}
\usepackage{algpseudocode}
\usepackage{bbding}
\usepackage{multirow}
\usepackage{makecell}
\usepackage{textcomp }
 
\usepackage{amssymb }
\usepackage{amsmath}
\usepackage{CJKutf8}
\usepackage{url}
\usepackage[normalem]{ulem}
\usepackage{subfigure}
\usepackage{multicol}
\usepackage{multirow}
\useunder{\uline}{\ul}{}

\newcommand{\nop}[1]{}
\usepackage{CJKutf8}
\usepackage[utf8]{inputenc}

\usepackage{color}

\setcopyright{acmcopyright}
\copyrightyear{2023}
\acmYear{2023}
\setcopyright{acmlicensed}\acmConference[KDD '23]{Proceedings of the 29th ACM SIGKDD Conference on Knowledge Discovery and Data Mining}{August 6--10, 2023}{Long Beach, CA, USA}
\acmBooktitle{Proceedings of the 29th ACM SIGKDD Conference on Knowledge Discovery and Data Mining (KDD '23), August 6--10, 2023, Long Beach, CA, USA}
\acmPrice{15.00}
\acmISBN{979-8-4007-0103-0/23/08}
\acmDOI{10.1145/3580305.3599862}

\begin{document}

\renewcommand{\algorithmicrequire}{\textbf{Input:}}  
\renewcommand{\algorithmicensure}{\textbf{Output:}} 

\title{M3PT: A Multi-Modal Model for POI Tagging}

\author{Jingsong Yang}
\affiliation{%
  \institution{School of Data Science, Shanghai Key Laboratory of Data Science,\\ Fudan University}
  \city{Shanghai}
  \country{China}
}
\email{jsyang21@m.fudan.edu.cn} 

\author{Guanzhou Han}
\affiliation{%
  \institution{Alibaba Group}
  \city{Hangzhou}
  \country{China}
}
\email{gavin.hgz@alibaba-inc.com}

\author{Deqing Yang}
\authornote{The two corresponding authors have the same contribution.}
\affiliation{%
  \institution{School of Data Science, Shanghai Key Laboratory of Data Science,\\ Fudan University}
  \city{Shanghai}
  \country{China}
}
\email{yangdeqing@fudan.edu.cn}

\author{Jingping Liu}
\authornotemark[1]
\affiliation{%
  \institution{School of Information Science and Engineering, East China University of Science and Technology}
  \city{Shanghai}
  \country{China}
}
\email{jingpingliu@ecust.edu.cn}

\author{Yanghua Xiao}
\affiliation{%
  \institution{School of Computer Science, Shanghai Key Laboratory of Data Science, Fudan University}
  \city{Shanghai}
  \country{China}
}
\email{shawyh@fudan.edu.cn}

\author{Xiang Xu}
\affiliation{%
  \institution{Alibaba Group}
  \city{Hangzhou}
  \country{China}
}
\email{alexander.xx@alibaba-inc.com}

\author{Baohua Wu}
\affiliation{%
  \institution{Alibaba Group}
  \city{Hangzhou}
  \country{China}
}
\email{zhengmao.wbh@alibaba-inc.com}

\author{Shenghua Ni}
\affiliation{%
  \institution{Alibaba Group}
  \city{Hangzhou}
  \country{China}
}
\email{shenghua.nish@alibaba-inc.com}

\renewcommand{\shortauthors}{Yang and Han,et al.}

\begin{abstract}
    \input{000abs_correct.tex}
\end{abstract}


\maketitle
\vspace{-0.2cm}
\section{Introduction}
\label{sec:intro}
\input{010intro_correct.tex}

\section{Related Work}
\label{sec:related}
\input{020related}
\label{sec:method}

\input{031over_correct.tex}

\input{032meth_correct.tex}

\input{042exp_correct.tex}


\section{Conclusion}
\label{sec:conclusion}
\input{060con_correct.tex}

\begin{acks}
\label{sec:acknowledgements}
\input{070acknowledgements.tex}

\end{acks}

\newpage
\bibliographystyle{ACM-Reference-Format}
\bibliography{main.bib}

\end{document}

%% file: 000abs_correct.tex
POI tagging aims to annotate a point of interest (POI) with some informative tags, which facilitates many services related to POIs, including search, recommendation, and so on. 
Most of the existing solutions neglect the significance of POI images and seldom fuse the textual and visual features of POIs, resulting in suboptimal tagging performance. In this paper, we propose a novel \textbf{M}ulti-\textbf{M}odal \textbf{M}odel for \textbf{P}OI \textbf{T}agging, namely \textbf{M3PT}, which achieves enhanced POI tagging through fusing the target POI's textual and visual features, and the precise matching between the multi-modal representations. Specifically, we first devise a domain-adaptive image encoder (DIE) to obtain the image embeddings aligned to their gold tags' semantics. Then, in M3PT's text-image fusion module (TIF), the textual and visual representations are fully fused into the POIs' content embeddings for the subsequent matching. In addition, we adopt a contrastive learning strategy to further bridge the gap between the representations of different modalities
. To evaluate the tagging models' performance, we have constructed two high-quality POI tagging datasets from the real-world business scenario of \emph{Ali Fliggy}. Upon the datasets, we conducted the extensive experiments to demonstrate our model's advantage over the baselines of uni-modality and multi-modality, and verify the effectiveness of important components in M3PT, including DIE, TIF and the contrastive learning strategy.

%% file: 010intro_correct.tex
A point of interest (POI) is a specific location that someones may feel helpful or interesting, including a park, restaurant, shop, museum, and so on. In the last decades, various services related to POIs have become very popular on Web. POI tagging, i.e., annotating POIs with some informative tags (labels), which not only help users better understand POIs' characteristics, but also are useful to discover the relatedness or similarities between different POIs. As a result, POI tagging can facilitate many downstream applications, such as POI search \cite{GaoLLXZ22} and recommendation \cite{KammererNPC09, YanCLNLX23}. 

\begin{CJK}{UTF8}{gbsn}
Many previous solutions achieve POI tagging based on \textbf{textual data} but only have limited performance. These solutions generally extract some features from the textual data relevant to a given POI, to infer the probability that each tag can be used to annotate the POI. The textual data mainly includes users' check-in logs \cite{KrummR13, KrummRC15}, POI taxonomy \cite{WangQPZX17} and descriptions \cite{LagosAC20}. However, considering only textual data for POI tagging may suffer from the problems of false positive (FP) and false negative (FN). FP refers to the fact that it might annotate the POI with the tags that are semantically related to the textual data but incorrect in fact. FN refers to the fact that it might overlook the tags that are semantically dissimilar to the textual data but correct in fact. 

\begin{figure}[t]
    \includegraphics[width=0.49\textwidth]{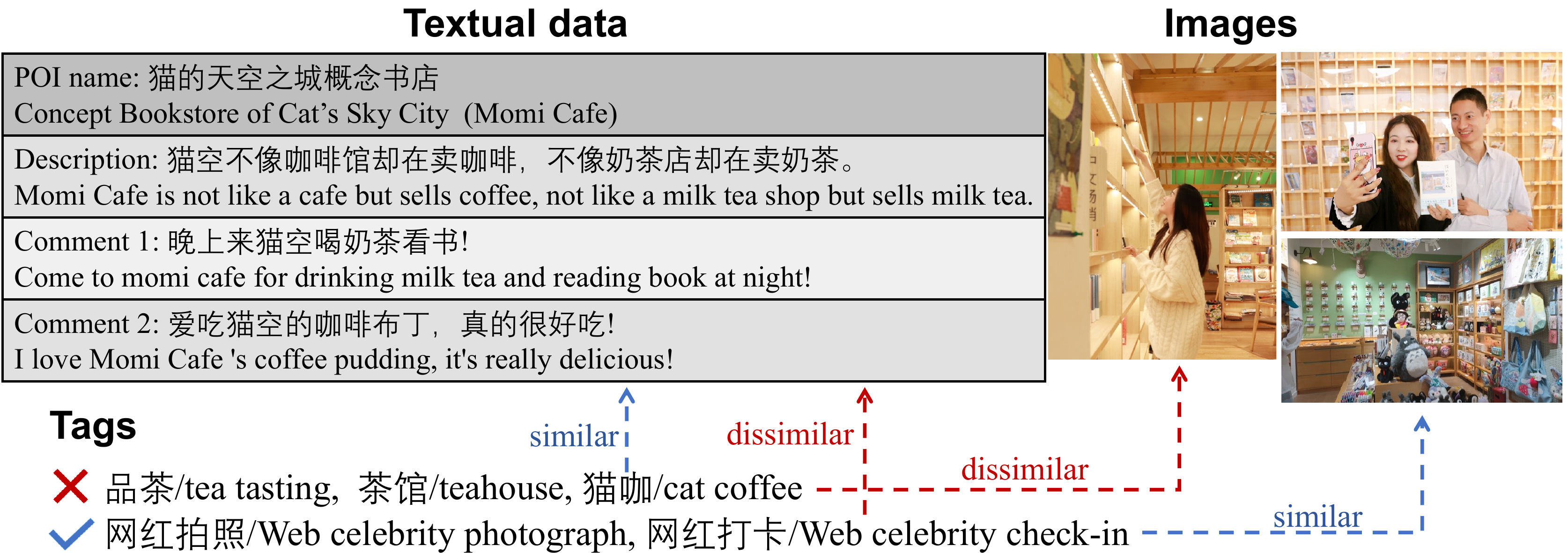}
    \vspace{-0.5cm}
    \caption{A toy example to demonstrate the effects of incorporating images into POI tagging.}
    \label{fig:intro}
    \vspace{-0.3cm}
\end{figure}

We have found that there is sufficient \textbf{visual data} (such as images) related to POIs on many real-world platforms, which is in fact an essential supplement of the textual data, to solve the aforementioned problems. For example, Fig. \ref{fig:intro} displays not only the textual data of a specific POI (Momi Cafe) including its name, description and user comments, but also some images posted in the comments. As shown in the figure, the FP and FN problems are alleviated or solved if the images are considered besides the textual data. Specifically, the POI was wrongly labeled with `品茶/tea tasting' due to the comment `晚上来猫空喝奶茶看书! / Come to Momi Xafe for drinking milk tea and reading book at night!'. While the images do not display the scene of tea tasting, hence helping the model correct such FP problem. Besides, the incorrect tag `茶馆/teahouse', and `猫咖/cat coffee' could be filtered out since their semantics are not related to the images. For FN, the correct tag `网红拍照/Web celebrity photograph' and `网红打卡/Web celebrity check-in' would be inferred as they have similar semantics to the images, although they are less semantically similar to the texts. The significance of POI images on discovering correct tags inspires us to leverage the textual and visual data simultaneously to achieve effective POI tagging.


\end{CJK}

A straightforward approach of leveraging images is to transfer an existing multi-modal model (e.g., CLIP \cite{RadfordKHRGASAM21} and BLIP \cite{0001LXH22}). However, this approach's effect is limited for the following reasons, especially given that the tagging task we focus on in this paper is specific to a real-world tour scenario of \emph{Ali Fliggy}\footnote{It is a famous Chinese tour platform and its URL is \url{https://www.fliggy.com}.}. 1) There is a gap between the textual and visual representations in general, which would likely lead to the mismatching between the POI and the gold tags. 2) Leveraging the existing image pre-trained models (such as ViT \cite{DosovitskiyB0WZ21}) might result in unsatisfactory representations of POI images, since our task is a domain-specific tagging task.

To address these problems, in this paper we propose a novel \textbf{M}ulti-\textbf{M}odal \textbf{M}odel for \textbf{P}OI \textbf{T}agging, namely \textbf{M3PT}, which is built based on a matching framework. Our model achieves enhanced tagging performance through the full fusion of the given POI's textual and visual features, and the precise matching between the POI's multi-modal representation and the candidate tag's representation. Specifically, in M3PT's feature encoding module, we devise a \textbf{D}omain-adaptive \textbf{I}mage \textbf{E}ncoder, namely \textbf{DIE}, to obtain the embeddings of POI images aligned to their gold tags' semantics. Then, we design a \textbf{T}ext-\textbf{I}mage \textbf{F}usion module, namely \textbf{TIF}, to fuse the textual and visual features of the POI. In TIF, we construct a clustering layer followed by an attention layer to distill the significant features for generating the POI's content embedding. Furthermore, we adopt a contrastive learning strategy to refine the embeddings of different modalities, and thus the cross-modality gap is bridged. In addition, the matching-based framework enables our M3PT to conveniently achieve the precise matching for a new tag, which is beyond the premise in traditional multi-label classification, i.e., the classification towards the close set of predefined labels. 

Our contributions in this paper are summarized as follows.

1. To the best of our knowledge, this is the first work to exploit a multi-modal model incorporating the textual and visual semantics to achieve the POI tagging on a real-world business platform. To this end, we propose a novel POI tagging model M3PT based on a multi-modal matching framework, to achieve the precise matching between POIs and tags.
    
2. We specially devise a domain-adaptive image encoder DIE in our model to obtain the optimal embedding for each input image, which is aligned to the semantics of the image's gold tags. The image embeddings generated by DIE are better adaptive to the requirements of the real-world scenario, resulting in enhanced tagging performance.

3. We have constructed two high-quality POI tagging datasets from the real-world tour scenario of Ali Fliggy, upon which we evaluate the models's tagging performance. Our extensive experiment results not only demonstrate our M3PT's advantage over the previous models of uni-modality and multi-modality, but also justify the effectiveness of the important components in M3PT.


%% file: 020related.tex
\subsection{POI Tagging Model}
Existing POI tagging solutions can be divided into three groups. The first group uses users' check-in data of POIs as inputs, and extracts discriminative features to predict location labels. Krumm et al. \cite{KrummR13, KrummRC15} presented a set of manually designed features that are extracted from the public place logs and individual visits. Some other methods \cite{HegdePH13} leverage the features of user check-in activities and other behavior data to train the generative probabilistic models to infer POI tags. The authors in \cite{YeSLYJ11, WangQPZX17, ZhuZL013, YangLC16} discussed more comprehensive check-in data including POI unique identifiers, user unique identifiers, the number, time and duration of check-ins, the latitude/longitude of user positions, as well as user demographic information. 
Label annotation for POIs was first studied in \cite{YeSLYJ11}, which introduces a collective classification method for feature extraction.
The authors in \cite{ZhuZL013} studied how to select the most relevant features for POI tagging. Yang et al. \cite{YangLC16} proposed an updatable sketching technique to learn compact sketches from user activity streams and then used a KNN classifier to infer POI tags. Wang et al. \cite{WangQPZX17} proposed a graph embedding method to learn POI embeddings from POI temporal bipartite graphs, which are then used by an SVM classifier of POIs labels.

\begin{figure*}[t]
    \includegraphics[width=0.98\textwidth]{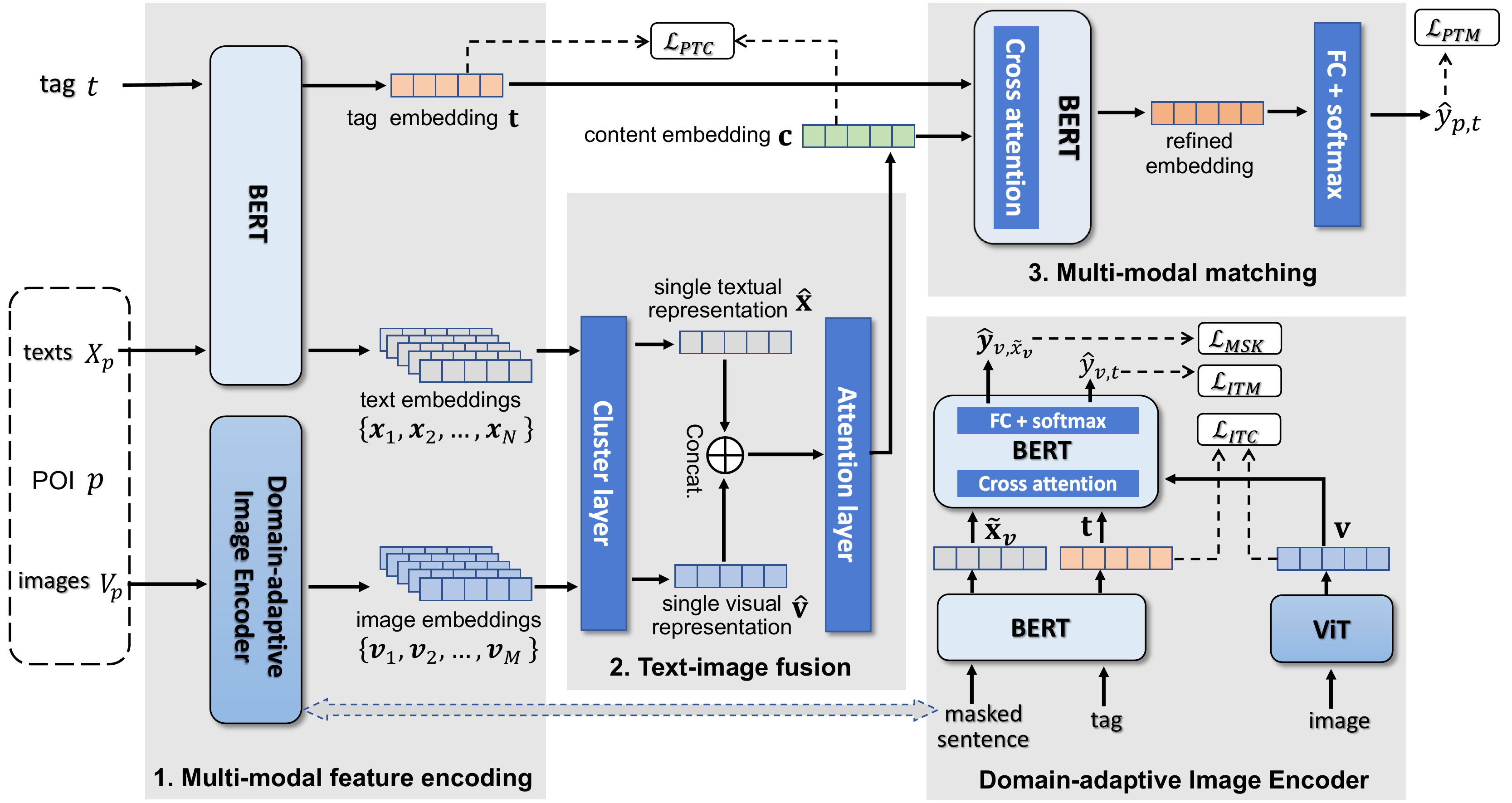}
    \vspace{-0.2cm}
    \caption{Our proposed M3PT consists of three modules. In the feature encoding module with a domain-adaptive image encoder (DIE), the textual and visual data of the given POI, and the candidate tag are encoded into the feature embeddings, respectively. Then, in the text-image fusion module (TIF), the text and image feature embeddings are fused into the POI's content embedding. At last, the final probability is computed in the multi-modal matching module based on the matching between the content embedding and tag embedding.}
    \label{fig:model}
    \vspace{-0.2cm}
\end{figure*}

The second group includes the methods proposed in \cite{HeYCZSL16,LagosAC20,ZhouGHZXJLX19}, which also use more fine-grained information of POIs besides user check-in data. For example, \cite{HeYCZSL16} includes general tags that may be related to categories and other information, e.g., "godzilla". \cite{ZhouGHZXJLX19} is the first to use POI name and address tokens, that are obtained by the pre-training on a domain and language-specific corpus. 
Lagos et al. \cite{LagosAC20} found that POIs have several distinctive properties, including notably multiscript names, geospatial identities, and temporally defined context. Thus they proposed an approach to complete POI semantic tags in a crowdsourced database automatically. 

Similar to our work, the third group leverages image tagging, 
of which many solutions use ViT \cite{DosovitskiyB0WZ21} as the backbone to accomplish multiple image classification. Although many algorithms have been proposed for automatic image annotation \cite{BarnardDFFBJ03, GohCL05, ChengZ0TL18}, image tag refinement is treated as an independent problem, and has become an exciting issue \cite{LinDH0Y13, FengFJJ14, LiSLZ16, ZhangWH17}. 

\vspace{-0.1cm}
\subsection{Vision-language Pre-trained Model}
Vision-language pre-training (VLP) aims to improve the performance of downstream vision and language tasks by pre-training models on large-scale image-text pairs. The pre-training tasks of VLP models mainly include image-text contrastive learning and language modeling (LM) based tasks. VirTex \cite{abs-2006-06666}, ICMLM \cite{abs-2008-01392}, and ConVIRT \cite{abs-2010-00747} have demonstrated the potential of Transformer-based LM, masked LM, and contrastive LM on learning image representations from the text. CLIP\cite{RadfordKHRGASAM21}, BLIP \cite{0001LXH22}, FILIP \cite{YaoHHLNXLLJX22}, ALIGNm \cite{JiaYXCPPLSLD21} and UNIMO \cite{LiGNXLL0020}  mainly make use of cross-modal contrastive learning which aligns the textual and visual information into a unified semantic space. VisualBERT \cite{abs-1908-03557}, UNITER \cite{ChenLYK0G0020}, and M6 \cite{abs-2103-00823} employ LM-like objectives, including both masked LM (e.g., Masked Language/Region Modeling) and auto-regressive LM (e.g., image captioning, text-grounded image generation). In addition, some methods (such as BriVL \cite{abs-2103-06561}) rely on a pre-trained object detection model, such as Faster-RCNN \cite{RenHGS15}, to extract regional image features offline. It requires extra labeled bounding-box data and makes the approach less scalable. Recent efforts such as SOHO \cite{HuangZH0FF21} and SimVLM \cite{WangYYDT022} try to alleviate this burden via visual dictionary or PrefixLM \cite{RaffelSRLNMZLL20}.

%% file: 031over_correct.tex
\section{OVERVIEW}
\subsection{Problem Formalization}
Given a POI $p$, the task objective of our model is to filter out some appropriate tags from all candidate tags, to characterize $p$. Expressly, in our scenario of Ali Fliggy, some textual contexts and images related to $p$ are also provided to the model to achieve the tagging task better. The textual contexts include the name, category, description, and user comments of $p$. The images include $p$'s main picture displayed on the platform and the pictures posted in user comments. 
Formally, we denote all textual contexts of $p$ as 
$X_p=\{x_1, x_2, \ldots, x_N\}$, and all related images as $V_p=\{v_1, v_2, \ldots, v_M\}$. The model should compute the following matching score (probability) between $p$ and a candidate tag $t$, 
$$
\hat{y}_{p,t}= \mathop{M3PT}(X_p,V_p,t),
$$
where $\mathop{M3PT}( )$ represents all operations in our model. With $\hat{y}_{p,t}$ and a well-chosen threshold $\pi$, the model predicts whether $t$ should be used to annotate $p$.

\subsection{Solution Framework}

As shown in Fig. \ref{fig:model}, our proposed M3PT achieving POI tagging pipeline can be divided into the following three phases.

\paragraph{1. Feature Encoding} The candidate tag $t$, the textual contexts, and images of the given POI $p$ are encoded into the feature embeddings in this first phase. The relevant operations are conducted in the feature encoding module of our model, which consists of a text encoder and an image encoder. Specifically, we take BERT$_{\text {base}}$ \cite{devlin2018bert} as the text encoder to encode  $t$ into a tag embedding and encode $X_p$ into a set of text feature embeddings of $p$ simultaneously. For the image feature encoding of $p$, we adopt a new image encoder DIE, which is specific to our tagging task in Fliggy's domain, as we mentioned in Section \ref{sec:intro}. We will introduce the details of the feature encoding module, including DIE, in Section \ref{sec:encode}.

\paragraph{2. Text-image Fusion} The operations in the second phase are conducted in the TIF module of our model. The TIF mainly consists of two layers. The first layer is a clustering layer, through which $p$'s text embeddings and image embeddings obtained from the first phase are aggregated into a single textual representation and a single visual representation of $p$, respectively. Then, these two representations are concatenated and fed into the attention layer of TIF to generate $p$'s content embedding for the POI-tag matching in the subsequent phase. As we emphasized in Section \ref{sec:intro}, the clustering and attention operations in TIF 
are proposed to distill the significant features to represent $p$'s content, whose details will be introduced in Section \ref{sec:TIF}.

\paragraph{3. Multi-modal Matching}
To conduct the operations in the third phase, we build a multi-modal matching module fed with $p$'s content embedding and $t$'s tag embedding generated in the previous phases to output the final score $\hat{y}_{p,t}$. The main body of this module is also a BERT$_{\text {base}}$ encoder, whose details will be introduced in Section \ref{sec:match}.

%% file: 032meth_correct.tex
\section{METHODOLOGY}

\subsection{Multi-modal Feature Encoding}\label{sec:encode}

In the feature encoding module of our model, the multi-modal features of $p$ (including textual and visual features) and $t$'s textual features are first encoded into embeddings, respectively.

Specifically, $p$'s textual inputs $\{x_1, x_2, \ldots, x_N\}$ are fed into a text encoder constructed based on BERT$_{base}$ to generate a set of $p$'s text embeddings. The text encoder is initialized with the first 6 layers of BERT$_{base}$. Formally, for an input text $x_i (1\leq i\leq N)$, we get its feature embedding as
\begin{equation}
\mathbf{x}_i=\mathop{EN_T}(x_i)\in\mathbb{R}^D, 
\end{equation}
where $\mathop{EN_T}()$ denotes the encoding operations in the text encoder, and $D$ is the embedding's dimension. 

Simultaneously, $t$ is also fed into the same text encoder to obtain its tag embedding as
\begin{equation}
\mathbf{t}=\mathop{EN_T}(t)\in\mathbb{R}^D.
\end{equation}

\subsubsection{Domain-adaptive Image Encoder}
Given that our tagging task is specific to the real-world scenario of Fliggy, we propose a new image encoder DIE to encode the input images of $p$, i.e., $\{v_1, v_2, \ldots, v_M\}$, into a set of image feature embeddings, instead of using an existing image pre-trained model \cite{abs-1908-03557, DosovitskiyB0WZ21}. To adapt to the particular goal of our POI tagging task, the embedding of an image in our model should be learned concerning the tags with the same semantics as it, i.e., its gold tags. It indicates that the learned embedding of an image should help the model recognize its gold tags. To this end, unlike the traditional image encoders such as ViT \cite{DosovitskiyB0WZ21}, which only receive images as input, our DIE takes an image and the text having the same semantics to constitute an input sample. Since DIE is also a pre-trained encoder, we propose a pretraining mask task. 

\paragraph{Mask Learning}
Specifically, given an image $v$, we first compose a sentence to indicate $v$'s semantics, denoted as $x_v$, which includes a gold tag of $v$, denoted as $t_v$. Then, we convert $x_v$ into $\Tilde{x}_v$ by replacing one token in $t_v$ with a special token [MASK]. For example, if an image $v$ and its annotated tag $t_v$=`cup' are obtained, we have
$$
x_v=\text{``This is a cup''}, \quad \Tilde{x}_v=\text{``This is a [MASK]''}.
$$

Accordingly, we take $(v,\Tilde{x}_v)$ as the input of the mask learning task. As displayed in Fig. \ref{fig:model}, $i$ and $\Tilde{x}_v$ are fed into a ViT encoder and a BERT$_{base}$ in DIE, to generate $v$'s embedding $\mathbf{v}\in\mathbb{R}^D$ and $\Tilde{x}_v$'s embedding $\Tilde{\mathbf{x}}_v\in\mathbb{R}^D$, respectively. Then, based on these two embeddings, DIE computes the following probability distribution indicating which token is on the position of [MASK] as,
\begin{equation}\label{eq:pre}
\hat{\mathbf{y}}_{v,\Tilde{x}_v}=\mathop{Pre}(\mathbf{v},\Tilde{\mathbf{x}}_v),
\end{equation}
where $\mathop{Pre}()$ represents all calculations in the prediction layer of DIE. Specifically, $\hat{\mathbf{y}}_{v,\Tilde{x}_v}$ is a vector of $U$ dimensions and $U$ is the size of token vocabulary. Each entry of $\hat{\mathbf{y}}_{v,\Tilde{x}_v}$ is the probability that the corresponding token is on the position of [MASK]. To compute this probability distribution more precisely, $\mathbf{v}$ is first refined in terms of $\Tilde{\mathbf{x}}_v$ through the cross-attention operations as \cite{vaswani2017attention}. Then $\hat{\mathbf{y}}_{v,\Tilde{x}_v}$ is computed based on the refined $\mathbf{v}$ through a fully-connected layer and softmax operation. At last, a token with the largest probability in $\hat{\mathbf{y}}_{v,\Tilde{x}_v}$ is predicted as on the position of [MASK].

To pretrain DIE, we formulate the mask learning's loss as
\begin{equation}
\mathcal{L}_{MSK}=\sum\limits_{(v,u)\in\mathcal{N}_{MSK}} \mathop{CE}(\mathbf{y}_{v,\Tilde{x}_v},\hat{\mathbf{y}}_{v,\Tilde{x}_v}),
\end{equation}
where $\mathop{CE}()$ is the cross-entropy function and $\mathcal{N}_{MSK}$ is the training set. $\mathbf{y}_{v,\Tilde{x}_v}\in\mathbb{R}^U$ is the indicator vector, in which the entries of the real tokens are 0 and the rest entries are 0.

\paragraph{Contrastive Learning}
In addition, to better align $v$'s embedding to its gold tags' embeddings, we further consider a contrastive learning loss. We first define $v$'s similarity score to each paired tag $t$ as
\begin{equation}\label{eq:s}
    s(v,t)= \frac{\exp \big(\mathbf{v}^\top\mathbf{t}/ \tau\big)}{\sum\limits_{t'\in \mathcal{T}_v} \exp \big(\mathbf{v}^\top\mathbf{t}'/ \tau\big)},
\end{equation}
where $t'$ is any one tag in $v$'s paired tag set $\mathcal{T}_v$ including its gold tag $t_v$, and $\mathbf{t}\in\mathbb{R}^D$ is $t$'s embedding obtained from the BERT$_{base}$ encoder. In addition, $\tau$ is the temperature used to maintain the balance between the alignment and consistency of the contrasted samples \cite{WangL21a}. To train the model better in contrastive learning, we tend to select the incorrect tags but semantically similar (having close embeddings) to the image as hard negative samples. Accordingly, a small $tau$ makes the model focus on discriminating the hard negative samples, resulting in better performance.

Similarly, a tag $t$'s similarity score to each paired image $v$ is 
\begin{equation}
    s(t,v)= \frac{\exp \big(\mathbf{t}^\top\mathbf{v}/ \tau\big)}{\sum\limits_{v'\in \mathcal{V}_t} \exp \big(\mathbf{t}^\top\mathbf{v}'/ \tau\big)},
\end{equation}
where $v'$ is any one image in $t$'s paired image set $\mathcal{V}_t$ including the gold images with the same semantics as $t$. Then, the contrastive learning loss is
\begin{equation}\label{eq:Litc}
\mathcal{L}_{ITC}=\frac{1}{2}\bigg\{\sum\limits_{(v,t)\in\mathcal{N}_{ITC}}\mathop{CE}\big(y_{v,t},s(v,t)\big)+ \sum\limits_{(t,v)\in\mathcal{N}_{ITC}}\mathop{CE}\big(y_{t,v},s(t,v)\big)\bigg\},
\end{equation}
where $y_{v,t}$ and $y_{t,v}$ are both the real label indicating the matching between $v$ and $t$.

\paragraph{Image-tag Matching Learning}
Furthermore, to better achieve the matching between images and tags, we append an image-tag matching loss for DIE's pretraining. We use $\hat{y}_{v,t}$ to denote the matching probability between $v$ and $t$. As the operations in Equation \ref{eq:pre}, $\mathbf{v}$ is first refined in terms of $\mathbf{t}$ through cross-attention and then used to compute $\hat{y}_{v,t}$. Then, the matching loss is
 \begin{equation}\label{eq:Litm}
\mathcal{L}_{ITM}=\sum\limits_{(v,t)\in\mathcal{N}_{ITM}} \mathop{CE}(y_{v,t},\hat{y}_{v,t}).
\end{equation}

 \bigskip
 After all, the overall loss of DIE's pre-training is\footnote{The same coefficient for each sub-loss is the best choice based on our tuning studies.}
\begin{equation}\label{eq:Ldie}
\mathcal{L}_{DIE}=\mathcal{L}_{MSK}+\mathcal{L}_{ITC}+\mathcal{L}_{ITM}.
\end{equation}

With the pre-trained DIE, the feature embedding of an image $v_i (1\leq i\leq M)$ is obtained by
\begin{equation}\label{eq:v}
\mathbf{v}_i=\mathop{DIE}(v_i)\in\mathbb{R}^D, 
\end{equation}
where $\mathop{DIE}()$ denotes the encoding operations in DIE. 

\subsection{Text-image Fusion}\label{sec:TIF}
The operations in this phase are conducted in the TIF module, where the textual and visual embeddings of $p$ are fused into a condensed representation, namely content embedding, to achieve the subsequent matching conveniently. 

\subsubsection{Clustering Layer}
The first step in TIF is to aggregate the multiple embeddings from either modality into a single embedding. To this end, we first build a clustering layer in TIF to perform the following operations. 

For an image $v$'s feature embedding $\textbf{v}=[v_1, v_2, \ldots, v_D]$ obtained by Equation \ref{eq:v}, each $v_i (1\leq i\leq D)$ can be regarded as a frame-level descriptor of $v$. We used a clustering algorithm the same as NeXtVLAD \cite{lin2018nextvlad} to cluster the $d$ frame-level descriptors of $v$ into $K$ groups. Suppose the $k$-th cluster's centroid is $c_k (1\leq k\leq K)$, we refine $v_i$ in terms of $c_k$ as
\begin{equation}\label{eq:vi}
v^k_i=\alpha_k\left(\mathbf{v}\right)\left(v_i-c_k\right),
\end{equation}
where $\alpha_k(\mathbf{v})$ is a function measuring $\mathbf{v}$'s proximity to the $k$-th cluster. We adopt a fully-connected layer with softmax activation to obtain $\alpha_k(\mathbf{v})$ as 
\begin{equation}
\alpha_k(\mathbf{v})=\frac{e^{\mathbf{w}_k^\top \mathbf{v}+b_k}}{\sum_{k'=1}^K e^{\mathbf{w}_{k'}^\top\mathbf{v}+b_{k'}}},
\end{equation}
where $\mathbf{w}\in \mathbb{R}^D$ and $b$ are both trainable parameters. Furthermore, there are $K$ groups of $\mathbf{w},b$ in total.

Then, to aggregate the refined frame-level descriptors of all $M$ image embeddings, we get the $i$-th aggregated descriptor in terms of the $k$-th cluster through the following sum pooling,
\begin{equation}
    \hat{v}_i^k = \Sigma_{j=1}^M v_i^{j,k}, 1\leq i\leq D, 1\leq k\leq K,
\end{equation}
where $v_i^{j,k}$ is the $i$-th refined frame-level descriptor of the $j$-th image embedding obtained by Equation \ref{eq:vi}. Accordingly, we have $D\times K$ aggregated descriptors in total, which are then reduced into a single visual representation of $H$ dimensions, denoted as $\hat{\mathbf{v}}\in\mathbb{R}^H$, through a fully-connected layer.

To aggregate the $N$ text embeddings generated in the previous phase into a single textual representation, denoted as $\hat{\mathbf{x}}\in\mathbb{R}^H$, we adopt the same operations introduced above for generating the single visual representation.  

\subsubsection{Attention Layer}
Next, we input the concatenation of $\hat{\mathbf{x}}$ and $\hat{\mathbf{v}}$ into the attention layer in TIF. This layer reassigns the feature weights in the input through the attention and reshape operations and outputs an embedding of $D$ dimensions. Thus, the more significant features are highlighted in the output, which is just $p$'s content embedding, denoted as $\mathbf{c}\in \mathbb{R}^{D}$. Overall, $\mathbf{c}$ condenses the significant multi-modal features of $p$.

\subsection{Mutil-modal Matching}\label{sec:match}
The objective of the multi-model matching module in the third phase is to achieve the precise matching between $p$ and $t$, based on $p$'s content embedding $\mathbf{c}$ and $t$'s tag embedding $\mathbf{t}$. 

In fact, the architecture of this module is similar to the prediction layer in DIE since both of their objectives are to achieve the matching between two objects. Specifically, the candidate tag's embedding $\mathbf{t}$ is refined in terms of $p$'s content embedding $\mathbf{c}$ through the cross-attention operations and then used to compute $\hat{y}_{p,t}$ through a fully-connected layer and softmax operations. As last, M3PT predicts that $t$ is a correct tag of $p$ if $\hat{y}_{p,t}> \pi$.

\subsection{Model Training}
Similar to Equation \ref{eq:Litm}, the main loss of our model training is the following POI-tag matching loss,
 \begin{equation}
\mathcal{L}_{PTM}=\sum\limits_{(p,t)\in\mathcal{N}_{PTM}} \mathop{CE}(y_{p,t},\hat{y}_{p,t}),
\end{equation}
where $\mathcal{N}_{PTM}$ is the training set consisting of POI-tag pairs.

To improve the alignment of $p$'s content embedding and $t$'s tag embedding, we also append a contrastive learning loss the same as Equation \ref{eq:s}$\sim$\ref{eq:Litc} for our model training, except that $\mathbf{v}$ is replaced with $\mathbf{c}$ when computing $s(p,t)$ and $s(t,p)$. Thus, this contrastive learning loss is 
\begin{equation}\label{eq:Lptc}
\mathcal{L}_{PTC}=\frac{1}{2}\bigg\{\sum\limits_{(p,t)\in\mathcal{N}_{PTC}}\mathop{CE}\big(y_{p,t},s(p,t)\big)+ \sum\limits_{(p,t)\in\mathcal{N}_{PTC}}\mathop{CE}\big(y_{t,p},s(t,p)\big)\bigg\}.
\end{equation}
And the overall loss of M3PT's training is
\begin{equation}\label{eq:L}
\mathcal{L}=\mathcal{L}_{PTM}+\alpha\mathcal{L}_{PTC},
\end{equation}
where $\alpha$ is a controlling parameter.

%% file: 042exp_correct.tex
\section{Experiments}

\subsection{Dataset Construction}
We have constructed the following two datasets for our evaluation experiments, of which the detailed statistics are listed in Table \ref{tab:task-statistic}.

\begin{table}[t]
\caption{Statistics of our datasets.}\label{tab:task-statistic}
\vspace{-0.2cm}
\scalebox{0.9}
{
\begin{tabular}{lllllll}
\toprule
Dataset & POI \# & tag \# & POI-tag \# & tag \# & image \# & avg. text \#  \\ \
name & &&&per POI & per POI & per POI\\
\hline
MPTD1  &63,415 & 354 & 197,254& 3 &  64 & 126 \\
MPTD2  &6,415 & 286 & 27,486 & 4  &  8 & 16  \\ 
\bottomrule
\end{tabular}
\vspace{-0.2cm}
}
\end{table}

\noindent \textbf{MPTD1}:
The first dataset, named MPTD1 (Multi-modal POI Tagging Dataset), was constructed directly from the real-world tour scenario of Fliggy, including more than 60,000 POIs, 354 unique tags, and more than 190,000 POI-tag pairs. For each POI in this dataset, its related textual contexts (texts in short) include its full name, introduction, categories, and user comments collected from Fliggy's website. Each POI's related images include the main picture shown on the top of its introduction page, as well as the pictures posted in its user comments. We first collected the original tags for each POI in some ways, including basic rules, manual selections, semantic-based algorithms, etc. When sufficient original tags had been collected, the experts verified and refined them based on their actual effects on Fliggy’s tour platform. At last, the reserved tags were identified as the gold (gound-truth) tags for these POIs. All POIs were divided into the training set, validation set, and test set according to 8:1:1. In each set, besides the POIs, their gold tags, texts, and images were also included.

\noindent \textbf{MPTD2}:
Although the texts and images of each POI in MPTD1 are sufficient, the model training is time-consuming if we feed all of them into the model. In addition, most of them are not semantically related to the POI's tags. Thus, to compare all models' capabilities of leveraging a POI's textual and visual features that are semantically related to its gold tags more efficiently, we further constructed a more concise dataset MPTD2. We randomly selected about one-tenth of the POIs in MPTD1, together with their tags, texts, and images. For each selected POI, we recruited some volunteers to check its texts and images and only retain those with similar semantics to its tags. For example, only the comments that directly mention the tags or are highly semantically related to the tags were retained.
Similarly, only the images verified as semantically related to the POI's tags were retained. As a result, the texts and images retained in MPTD2 can be directly leveraged as pieces of evidence to judge the POI's matching to its tags. The ratio of the training set, validation set, and test set in MPTD2 is also 8:1:1.
\begin{table}[b]
\caption{Hyperparameter settings of M3PT.}
\vspace{-0.2cm}
\begin{tabular}{lcl}
\toprule
Notation & Value & Description \\ \midrule
  $\tau_1$ & 0.12 & temperature in $\mathcal{L}_{PTC}$\\
  $\tau_2$ & 0.08 & temperature in $\mathcal{L}_{ITC}$\\
  $\pi$ & 0.5 & threshold of prediction probability\\
   $D$ & 768 & embedding dimension\\
   $K$ & 64 & cluster number in TIF \\
   $H$ & 414 & textual/visual representation dimension in TIF \\ 
   $\alpha$ & 0.5 & controlling parameter in loss $\mathcal{L}$ (Equation \ref{eq:L})\\
  \bottomrule
\end{tabular}
\label{tab:hyperparameters}
\vspace{-0.2cm}
\end{table}

\begin{table*}[t]
\caption{Performance comparison results show M3PT's advantage over the baselines on the two datasets, where the best and second-best scores in each group are in bold and underlined, respectively.}
\label{tab:donwstream-result}
\vspace{-0.2cm}
    \centering
\resizebox{1.02\textwidth}{!}
{
    \begin{tabular}{|l|ccccccc|ccccccc|}
\Xhline{1pt}
          \multirow{2}*{\textbf{Model}} & \multicolumn{7}{c|}{\textbf{MPTD2}} &\multicolumn{7}{c|}{\textbf{MPTD1}} \\    \cline{2-15}
        ~ & M-F & M-R & M-F1 & P-e & R-e & F1-e & HLS & M-F & M-R & M-F1 & P-e & R-e & F1-e & HLS\\ \hline
        BERT & 74.67 & 26.65 & 39.26 & 75.72 & 29.51 & 42.40 & 0.0835 & 45.25 & 12.18 & 19.46 & 54.33 & 20.34 & 29.59 & 0.1285 \\ 
        ALBERT & 83.31 & \underline{29.53} & 44.16 & 85.14 & 34.26 & 48.85 & \underline{0.0627} & 54.23 & \underline{13.39} & 21.35 & 64.55 & 23.90 & 34.02 & \underline{0.0610} \\ 
        ERNIE & \underline{84.16} & 29.42 & \underline{45.56} & \underline{86.63} & \underline{36.06} & \underline{50.93} & 0.0683 & \underline{54.87} & 13.29 & \underline{21.93} & \underline{65.26} & \underline{23.97} & \underline{35.06} & 0.0857 \\ 
        M3PT(text) & \textbf{86.34} & \textbf{32.57 }& \textbf{47.11} & \textbf{88.38} & \textbf{38.63} & \textbf{53.72} & \textbf{0.0413} & \textbf{56.42} & \textbf{15.37} & \textbf{23.65} & \textbf{67.04} & \textbf{25.51} & \textbf{36.95} & \textbf{0.0512} \\ 
     
        Improv.\% & 2.59  & 10.29  & 3.40  & 2.02  & 7.13  & 5.48  & 34.13  & 2.82  & 14.79  & 7.84  & 2.73  & 6.42  & 5.39  & 16.07   \\
        \hline
        ResNet101 & 14.24 & \underline{9.40} & 11.32 & 18.15 & 10.68 & 13.44 & 0.1141 & 10.21 & 7.87 & 8.81 & 13.60 & 9.48 & 11.17 & 0.1364 \\ 
        ResNet101(ASL) & 14.37 & 9.36 & 11.36 & \underline{18.97} & 10.07 & 13.12 & 0.1163 & 10.04 & 8.07 & 8.94 & 13.51 & 9.22 & 10.96 & 0.1328 \\ 
        TResNet & 14.73 & 8.91 & 11.07 & 17.16 & 10.50 & 13.02 & \textbf{0.0915} & 10.80 & \underline{8.19} & 9.31 & 13.60 & 10.13 & 11.61 & \textbf{0.1243} \\ 
        ViT & 15.14 & 9.27 & 11.74 & 18.74 & 11.26 & \underline{14.06} & 0.1149 & 10.14 & \textbf{8.27} & 9.11 & 13.06 & 9.61 & 11.07 & 0.1470 \\ 
        ViT(Q2L) & \underline{15.26} & 9.39 & \underline{12.06} & 17.63 & \textbf{11.89} & 13.69 & 0.0952 & \underline{11.28} & 8.14 & \underline{9.45} & \underline{14.40} & \textbf{10.63} & \underline{12.23} & 0.1315 \\ 
        M3PT(image) & \textbf{15.85} & \textbf{10.95} & \textbf{12.93} & \textbf{19.34} & \underline{11.47} & \textbf{14.21} & \underline{0.0924} & \textbf{12.32} & 7.94 & \textbf{9.63} & \textbf{15.22} & \underline{10.42} & \textbf{12.37} & \underline{0.1293} \\   
        
        Improv.\% & 3.87  & 16.61  & 7.21  & 1.95  & -3.53  & 1.07  & -0.98  & 9.22  & -3.99  & 1.90  & 5.69  & -1.98  & 1.14  & -4.78 \\
        \hline
        M3TR & 86.49 & 31.64 & 46.52 & \underline{88.42} & \underline{39.03} & \underline{54.15} & 0.0620 & 57.41 & 15.34 & 24.29 & \underline{70.92} & 26.57 & 38.65 & 0.0784 \\
        TAILOR & \underline{87.72} & \underline{32.16} & \underline{46.83} & 87.34 & 37.80 & 52.76 & \underline{0.0427} & \underline{57.92} & \underline{16.26} & \underline{25.39} & 69.45 & \underline{27.66} & \underline{39.56} & \underline{0.0512} \\ 
        M3PT & \textbf{89.72} & \textbf{35.41} & \textbf{50.45} & \textbf{92.45} & \textbf{43.25} & \textbf{58.93} & \textbf{0.0242} & \textbf{59.51} & \textbf{17.56} & \textbf{27.11} & \textbf{72.15} & \textbf{28.50} & \textbf{40.85} & \textbf{0.0307}\\
        Improv.\% &2.28  & 10.11  & 7.73  & 4.56  & 10.81  & 8.83  & 43.33  & 2.75  & 8.00  & 6.77  & 1.73  & 3.04  & 3.26  & 40.04    \\
        \Xhline{1pt}
    \end{tabular}
}
\end{table*}

\subsection{Experimental Setup}
\subsubsection{Baselines}
We compared our M3PT with the following models in our experiments, which can be categorized into two groups of uni-modal models and one group of multi-modal models.

The baselines in the first group only leverage the textual features of POIs to achieve POI tagging. In these models, a given POI's texts are fed into their encoders to generate the textual embedding of the POI, which is used as the POI's content embedding. This group includes \noindent\textbf{BERT} \cite{devlin2018bert}, \textbf{ALBERT} \cite{lan2019albert} and  \textbf{ERNIE} \cite{ZhangHLJSL19}. The first two are both classic pre-trained language models. ERNIE incorporates knowledge maps into the pretraining task to improve its representation capability, which has been successfully employed in the previous multi-label classification task.

The baselines in the second group are also uni-modal models, which only leverage the POIs' images to achieve POI tagging. Specifically, only the images of a given POI are input into the models to generate the visual feature embedding used as its content embedding. This group includes the ResNet-based models including \textbf{ResNet-101} \cite{HeZRS16}, \textbf{ResNet101(ASL)} \cite{ben2020asymmetric} and \textbf{TResNet} \cite{RidnikLNBSF21}. All of them have been widely used to encode images in the computer vision (CV) community. In addition, we also considered \textbf{ViT} \cite{DosovitskiyB0WZ21} and \textbf{ViT-Q2L} \cite{abs-2107-10834}. Inspired by the success of Transformer \cite{vaswani2017attention} in natural language processing (NLP), these two models were built based on the architecture of self-attention and cross-attention, showing their advantages over the image encoders based on convolutional neural networks (CNNs).
 
As M3PT, the baselines in the third group are both multi-modal models, including \textbf{M3TR} \cite{1m3tr} and \textbf{TAILOR} \cite{zhang2022tailor}. M3TR employs the vision Transformer and builds a cross-modal attention module and a label-guided augmentation module to achieve multi-modal multi-label classification better. TAILOR first encodes the uni-modal features of texts and images and then fuses them into the cross-modal features as the encoder inputs to model the relationship between the modality and each label.

\subsubsection{Evaluation Metrics}
Our POI tagging task is just a multi-label classification if we regard the set of all tags as the label set of the POI classification. So we used the following metrics to evaluate all compared models' performance in our experiments. We first considered the label-based classification metric \textbf{M-P} (Macro Precision), \textbf{M-R} (Macro Recall) and \textbf{M-F1} (Macro F1) \cite{Yang99}. When computing their scores, we first took a tag along with all POIs annotated by it as one sample, and reported the average score of all samples. 
Second, we also considered the object-based classification metrics, including \textbf{P-e} (Precision-exam), \textbf{R-e} (Recall-exam), and \textbf{F1-e} (F1-exam) \cite{GodboleS04}. When computing three metrics' scores, we took a POI along with all of its tags as one sample, and reported the average score of all samples. In addition, we also considered \textbf{HLS} (Hamming Loss) \cite{GodboleS04}, which is used to measure the misclassification samples on a single label (tag). Thus, a more miniature HLS score indicates the model's better performance.

\subsubsection{Hyperparameter Settings}
We adopted AdamW \cite{LoshchilovH19} as the optimizer
Moreover, we set the initial learning rate to 1e-4 for pre-training. With steps increasing, we decreased the learning rate linearly to 1e-5. The settings of some crucial hyperparameters in our model are shown in Table \ref{tab:hyperparameters}, which were decided based on our parameter tuning studies. Specifically, we will display the results of tuning $\tau$ and $\pi$ afterward.

To reproduce our experiment results conveniently, we have published our model's source code on \url{https://github.com/DeqingYang/M3PT}.

\begin{table*}[t]
\caption{Ablation study results on the two datasets.}
 \label{tab:Ablation1}
 \vspace{-0.2cm}   
    \centering
\resizebox{1.02\textwidth}{!}
{
    \begin{tabular}{|lll|cc|cc|cc|cc|cc|cc|cc|}
\Xhline{1pt}

\multicolumn{17}{|c|}{\textbf{MPTD2}}\\ \hline
        PTC & TIF & Image &  M-P & Drop\%& M-R & Drop\%&M-F1 &Drop\%&P-e & Drop\%& R-e&Drop\% & F1-e& Drop\%&  HLS & Drop\%\\ 
        \hline
        \XSolidBrush & \XSolidBrush & \XSolidBrush& 80.71 & 10.04  & 26.03 & 26.49  & 39.36 & 21.98  & 81.1 & 12.28  & 36.06 & 16.62  & 49.92 & 15.29  & 0.0641  & 164.88   \\ 
        \XSolidBrush & \XSolidBrush & \Checkmark & 83.28 & 7.18  & 28.39 & 19.82  & 42.43 & 15.90  & 83.05 & 10.17  & 36.9 & 14.68  & 51.19 & 13.13  & 0.0630  & 160.33   \\ 
        \XSolidBrush & \Checkmark & \Checkmark & 85.91 & 4.25  & 30.9 & 12.74  & 45.45 & 9.91  & 86.55 & 6.38  & 37.04 & 14.36  & 51.87 & 11.98  & 0.0547  & 126.03   \\ 
        \Checkmark & \Checkmark & \XSolidBrush& 86.34 & 3.77  & 32.57 & 8.02  & 47.11 & 6.62  & 88.38 & 4.40  & 38.63 & 10.68  & 53.72 & 8.84  & 0.0413  & 70.66   \\ 
         \Checkmark & \XSolidBrush & \Checkmark & 87.81 & 2.13  & 32.36 & 8.61  & 47.25 & 6.34  & 89.16 & 3.56  & 38.05 & 12.02  & 53.37 & 9.43  & 0.0432  & 78.51   \\ 
        \Checkmark & \Checkmark & \Checkmark & \textbf{89.72} & - & \textbf{35.41} & - & \textbf{50.45} & - & \textbf{92.45} & - & \textbf{43.25} &- & \textbf{58.93} & - & \textbf{0.0242}  & - \\
\hline
 \multicolumn{17}{|c|}{\textbf{MPTD1}}\\ \hline
        PTC & TIF & Image &  M-P & Drop\%& M-R & Drop\%&M-F1 &Drop\%&P-e & Drop\%& R-e&Drop\% & F1-e& Drop\%&  HLS & Drop\%\\ 
        \hline

        \XSolidBrush & \XSolidBrush & \XSolidBrush& 49.67 & 16.54  & 13.81 & 21.36  & 21.61 & 20.29  & 56.13 & 22.20  & 20.7 & 27.37  & 30.07 & 26.39  & 0.0809  & 163.52   \\ 
        \XSolidBrush & \XSolidBrush & \Checkmark & 52.42 & 11.91  & 13.37 & 23.86  & 21.93 & 19.11  & 62.15 & 13.86  & 22.91 & 19.61  & 32.2 & 21.18  & 0.0753  & 145.28   \\ 
        \XSolidBrush & \Checkmark & \Checkmark & 54.62 & 8.22  & 14.41 & 17.94  & 22.8 & 15.90  & 65.26 & 9.55  & 24.05 & 15.61  & 35.67 & 12.68  & 0.0773  & 151.79   \\ 
        \Checkmark & \Checkmark & \XSolidBrush & 56.42 & 5.19  & 15.37 & 12.47  & 23.65 & 12.76  & 67.04 & 7.08  & 25.51 & 10.49  & 36.95 & 9.55  & 0.0542  & 76.55   \\ 
       \Checkmark & \XSolidBrush & \Checkmark & 57.79 & 2.89  & 16.34 & 6.95  & 25.58 & 5.64  & 69.55 & 3.60  & 25.9 & 9.12  & 37.02 & 9.38  & 0.0504  & 64.17   \\ 
        \Checkmark & \Checkmark & \Checkmark & \textbf{59.51} & - & \textbf{17.56} & -& \textbf{27.11} & -& \textbf{72.15} &- & \textbf{28.5} & - & \textbf{40.85} & - & \textbf{0.0307}  & - \\     
     \Xhline{1pt}
    \end{tabular}
}
\end{table*}

\subsection{Main Results}
The following performance scores of all compared models are reported as the average results of three runnings to alleviate the bias of single running. 
Table \ref{tab:donwstream-result} lists the overall tagging performance of all models on MPTD1 and MPTD2 datasets, where the best and second-best scores in each group are in bold and underlined, respectively. In particular, to verify our M3PT's capability of leveraging uni-modal features, we further proposed two ablated variants of M3PT. Wherein M3PT(text) only leverages the textual features, i.e., it only inputs the text embeddings into the cluster layer of TIF and directly uses the single textual representation $\hat{\mathbf{x}}$ as the POI's content embedding $\mathbf{c}$. Similarly, M3PT(image) only inputs image embeddings into TIF to generate the content embedding. M3PT's performance improvement ratio relative to the best baseline (underlined) is also listed in each group. The results demonstrate that M3PT performs the best on most metrics. Moreover, we also have the following observations and analyses.

1. Compared with the first group of textual modality and the third group of multi-modality, M3PT is not the best on some metrics. Through the investigation of the two datasets, we found that the images of a POI are semantically related to only about one-fifth of the POI's tags on average. It implies that M3PT can not exert its advantage sufficiently when only leveraging these images to discover the correct tags. 

2. Compared with MPTD1, M3PT's performance improvement over the baselines on MPTD2 is more apparent. It is because the POIs' texts and images in MPTD2 are more related to the tags, helping M3PT discover the matching across different modalities. Consequently, our model can achieve more precise POI tagging on MPTD2. It implies that our model is better at exploiting the high-quality dataset.

\subsection{Detailed Analysis}
\subsubsection{Ablation Studies}
We have conducted some ablation studies to justify the effects of the essential modules and strategy we designed in M3PT, including incorporating POI images, TIF, and the contrastive learning of POI-tag (PTC). Table \ref{tab:Ablation1} lists the results of different ablated variants of M3PT on MPTD2 and MPTD1, respectively, where the variant without image is just M3PT(text) in Table \ref{tab:donwstream-result}. In addition, in the variant without TIF, the single textual representation $\hat{\mathbf{x}}$ is just the sum of $N$ text embeddings. Similarly, the single visual representation $\hat{\mathbf{v}}$ is just the sum of $M$ image embeddings. Then, these two representations are concatenated and reshaped into $p$'s content embedding $\mathbf{c}$. 

Besides the metric scores, the performance drop ratios of all variants relative to M3PT are also listed in Table \ref{tab:Ablation1}. The results show that incorporating either one component of PTC, TIF, and image is helpful to M3PT's performance improvement. Compared with TIF and image, PTC is more helpful for M3PT to achieve precise POI tagging since the performance drop of the variant without PTC is the most obvious.

\begin{table}[t]
\caption{Performance comparisons of adopting different image encoder in M3PT.}
\vspace{-0.2cm}
    \begin{tabular}{lcccccc}
\toprule
          \textbf{Image} & \multicolumn{3}{c}{\textbf{MPTD2}} &\multicolumn{3}{c}{\textbf{MPTD1}} \\   \cmidrule(l){2-4} \cmidrule(l){5-7}
        \textbf{encoder} & mAP & M-F1 & F1-e& mAP & M-F1 & F1-e \\ \midrule
        
        ViT & 81.60 & 47.16 & \underline{56.93} & 66.20 & \underline{26.07} & 35.62\\ 
        CLIP & 81.36 & 46.93 & 55.40 & 66.94 & 25.70 & 36.15 \\ 
        BLIP & \underline{82.41} & \underline{47.31} & 55.03 & \underline{67.78} & 25.62 & \underline{37.49} \\ 
        DIE & \textbf{84.32} & \textbf{50.45 } & \textbf{58.93} & \textbf{69.14} & \textbf{27.11} & \textbf{40.85} \\        
 \bottomrule
    \end{tabular}
    \label{tab:image encoder}
\end{table}

\subsubsection{Impact of Domain-adaptive Image Encoder}

As mentioned before, the DIE specially designed in M3PT is an image encoder more adaptive to the requirements of the real-world scenario of Fliggy, resulting in M3PT's enhanced performance. To verify it, we further propose some variants of M3PT by replacing the DIE with some previous image encoders, including ViT \cite{DosovitskiyB0WZ21}, CLIP \cite{RadfordKHRGASAM21} and BLIP \cite{0001LXH22}.

\begin{figure}[t]
    \includegraphics[width=0.25\textwidth]{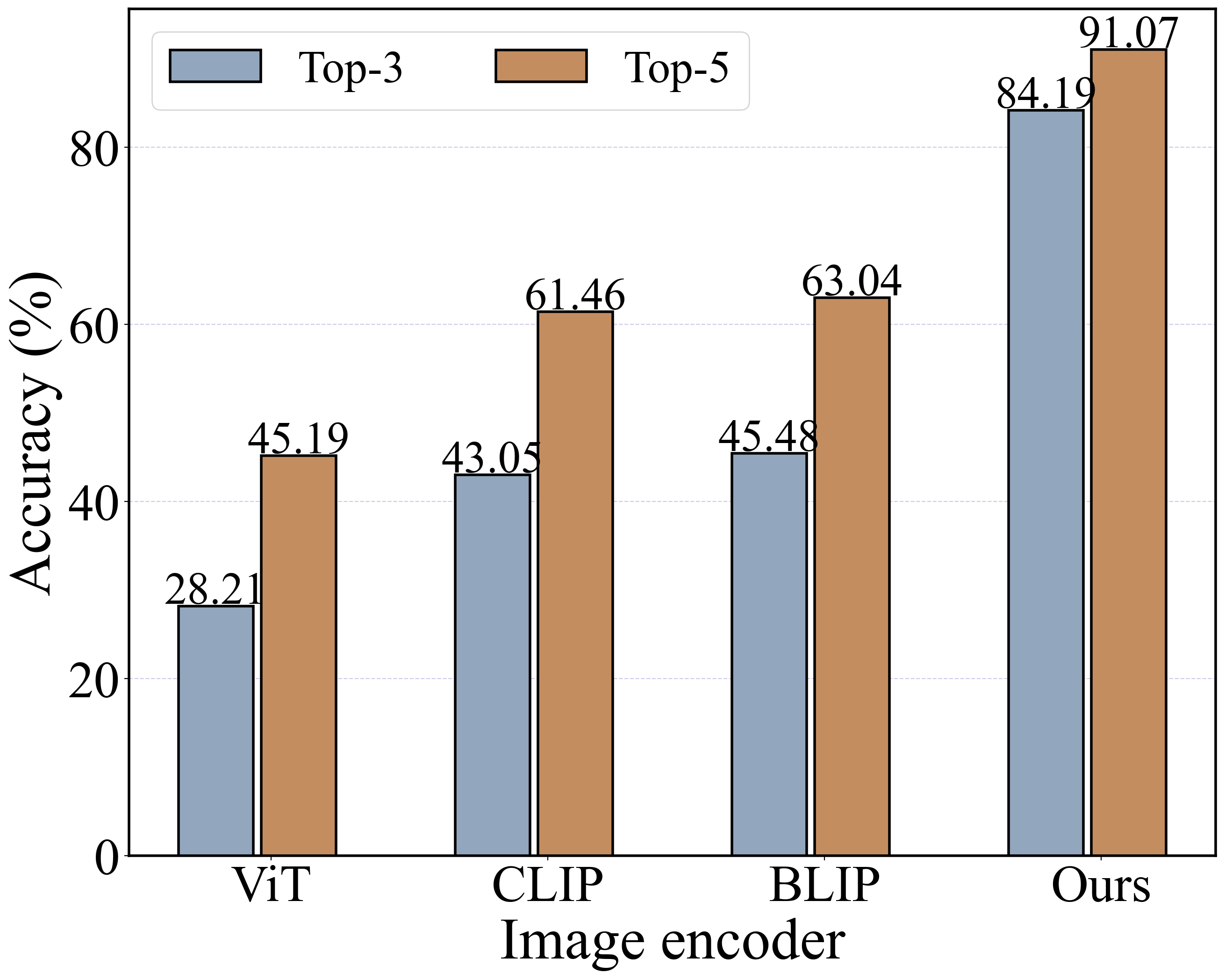}
    \vspace{-0.2cm}
    \caption{The precision of top-3/5 tags for the POIs on on MPTD2 predicted by the M3PTs with different image encoders.}
    \label{fig:top3/5}
    \vspace{-0.2cm}
\end{figure}

\begin{figure}[t]
    \includegraphics[width=0.48\textwidth]{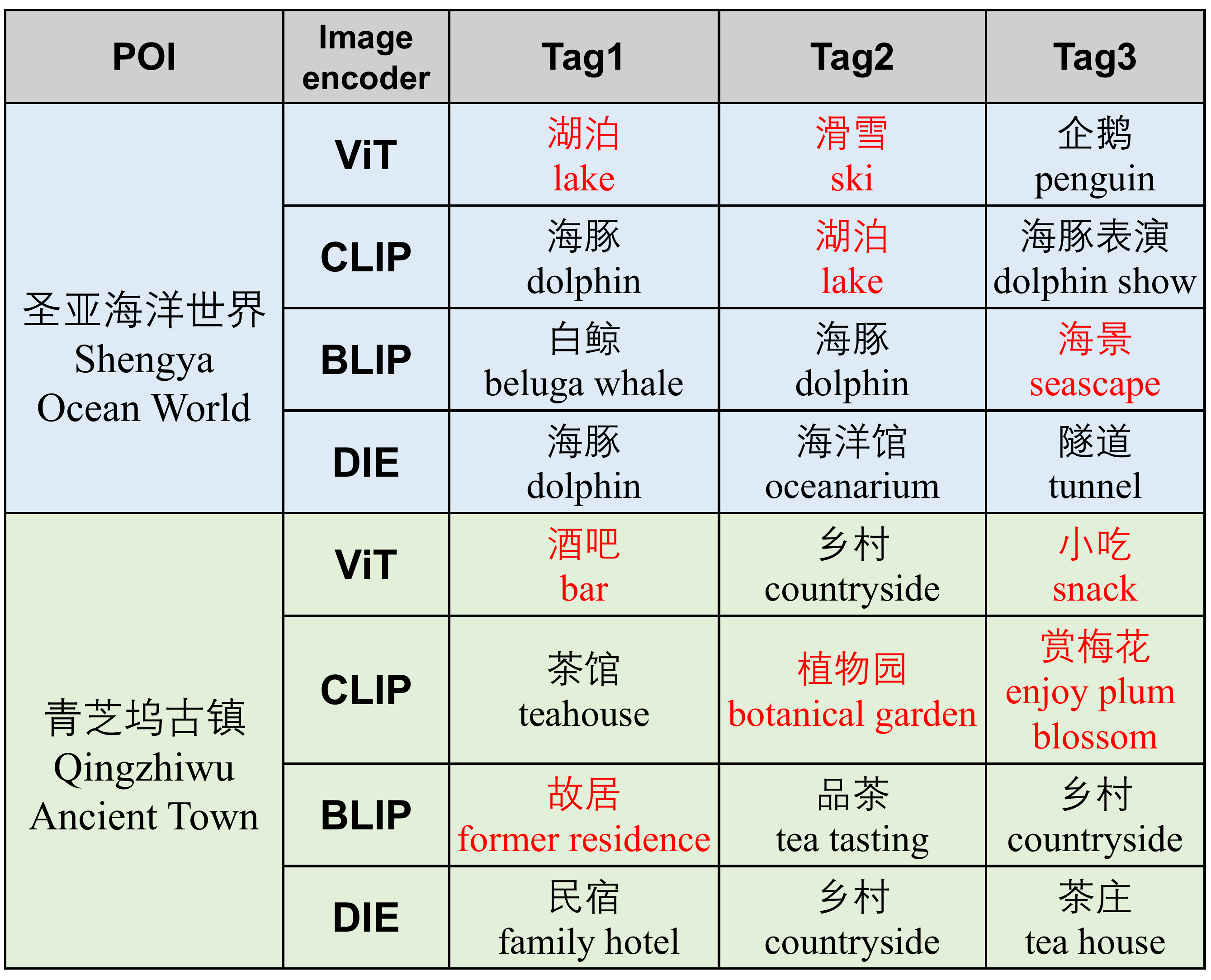}
    \vspace{-0.6cm}
    \caption{Top-3 tags predicted by the models for two POIs on MPTD2 (better viewed in color).}
    \label{fig:top3}
    \vspace{-0.2cm}
\end{figure}

Due to space limitation, we only report the compared models' scores of mAP, M-F1, and F1-e in Table \ref{tab:image encoder} since they are the most representative metric of the ranking metrics, label-based classification metrics, and object-based classification metrics. The results show that, with DIE, our M3PT performs better on both classification and ranking the correct tags. In addition, to concretely exhibit DIE's advantage in ranking the gold (correct) tags, in Fig. \ref{fig:top3/5}, we display the precision of top-3/5 tags predicted for the POIs on MPTD2. Obviously, all models' precision of top-5 is higher than that of top-3. Nevertheless, DIE's precision gap between top-3 and top-5 is much narrower than the other image encoders. Furthermore, in the table of Fig. \ref{fig:top3}, we list the top-3 tags predicted by the models for two POIs from MPTD2, where the incorrect tags are marked red. It also justifies that the M3PT with DIE can predict a bigger probability for the correct tags than the baselines.

\subsubsection{Performance on Ranking Correct Tags}
We also evaluated M3PT's performance on the three ranking metrics. However, due to space limitation, we only compared it with the baseline performing the best in each group according to Table \ref{tab:donwstream-result}. The comparison results are listed in Table \ref{tab:ranking metrics}, showing that our model still has an advantage in ranking the correct tags on higher positions.

\begin{table}[t]
\caption{Tag ranking performance comparisons between M3PT and the best baseline in each group.}\label{tab:ranking metrics}
\vspace{-0.2cm}
\resizebox{0.5\textwidth}{!}
{
\begin{tabular}{lcccccc}
\toprule
          \multirow{2}*{\textbf{Model}} & \multicolumn{3}{c}{\textbf{MPTD2}} &\multicolumn{3}{c}{\textbf{MPTD1}} \\   \cmidrule(l){2-4} \cmidrule(l){5-7}
        ~ & mAP & 1-Error & RL & mAP & 1-Error & RL \\ \midrule
               
        ERNIE & 60.05 & 0.3315 & 0.1620 & 78.21 & 0.2403 & 0.1461 \\
        M3PT(text) & \textbf{64.15} & \textbf{0.2251} & \textbf{0.1390} & \textbf{80.41} & \textbf{0.1528} & \textbf{0.1230} \\ \hline
        
        ViT(Q2L) & 13.90 & 0.6701 & 0.3902 & 18.90 & \textbf{0.2271} & 0.3610 \\
        M3PT(image) & \textbf{16.73} & \textbf{0.6278} & \textbf{0.3894} & \textbf{20.05} & 0.2315 & \textbf{0.3521} \\ 
        \midrule
        TAILOR & 63.28 & 0.3025 & 0.1242 & 81.50 & 0.2362 & 0.0936 \\ 
        M3PT & \textbf{69.14} & \textbf{0.2143} & \textbf{0.0827} & \textbf{84.32} & \textbf{0.1264} & \textbf{0.0541} \\ 
        \bottomrule
    \end{tabular}
}
\end{table}

\subsubsection{Hyperparameter Tuning}
For the important hyperparameters in M3PT, we have studied their impacts on the performance of POI tagging. Due to space limitation, we only display the results of $\tau_1$ and $\pi$.

\begin{figure}[t]
    \subfigure[MPTD2]{
       \centering
        \includegraphics[width=0.228\textwidth]{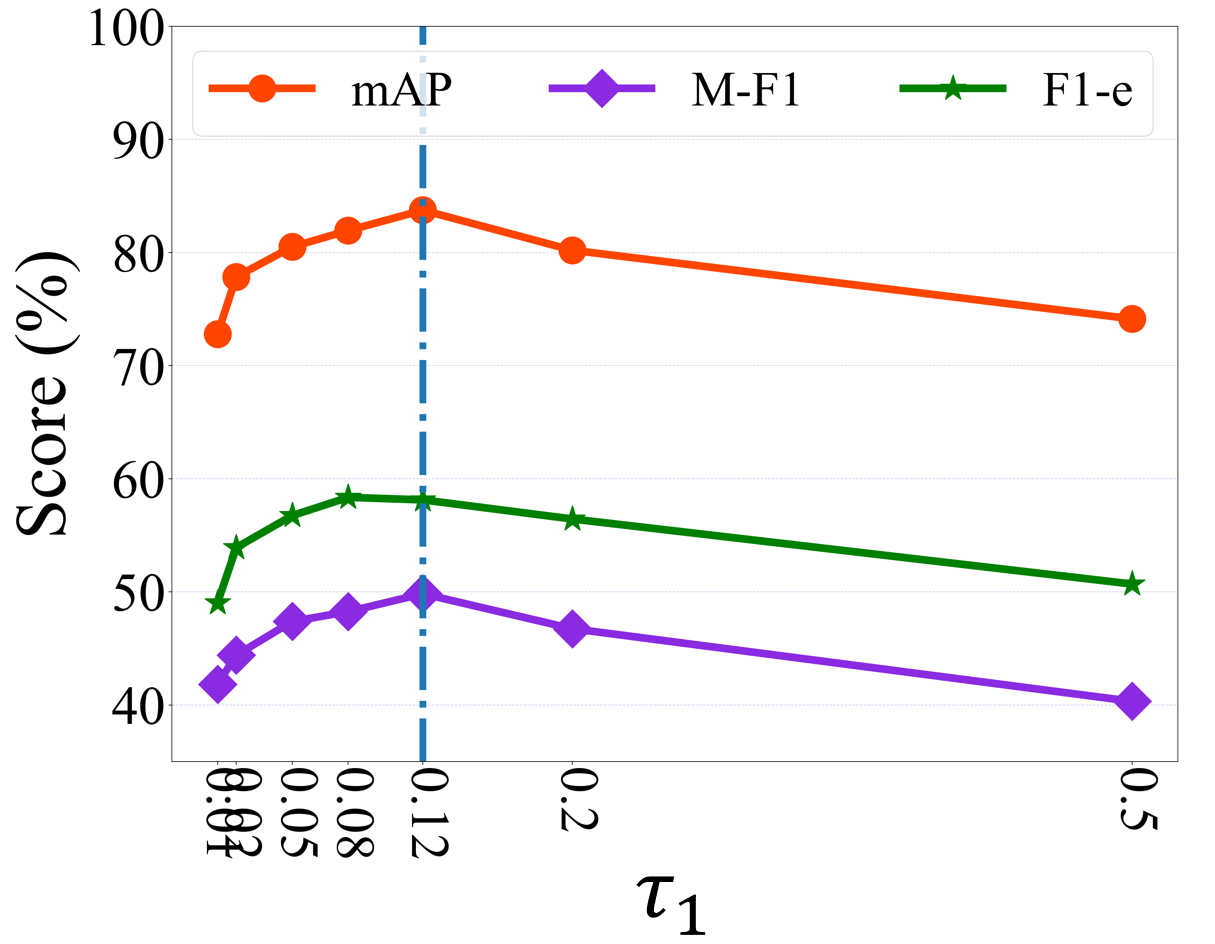}    
    }
    \hspace{-0.15cm}
    \subfigure[MPTD1]{
        \centering
        \includegraphics[width=0.228\textwidth]{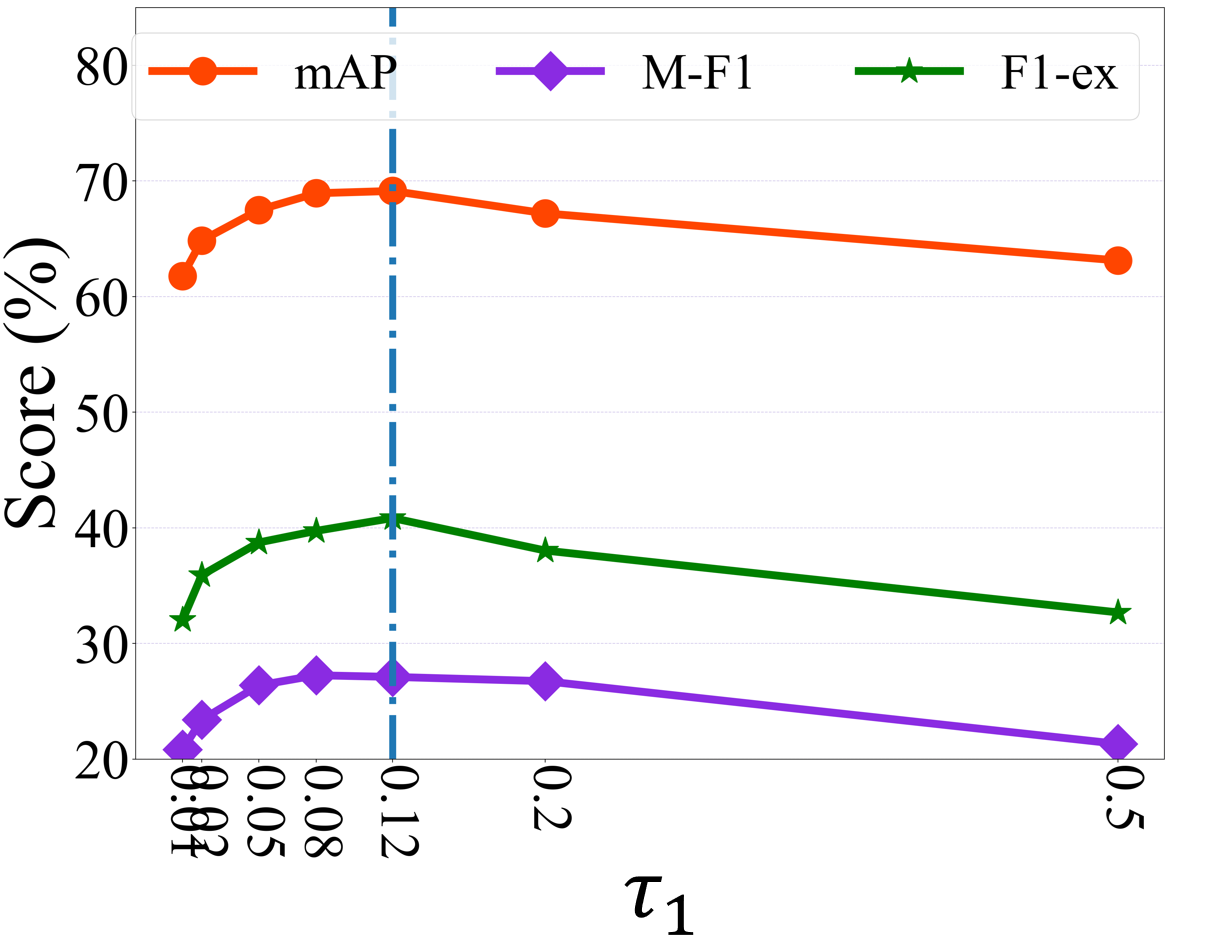}
    }
    \vspace{-0.2cm}
    \caption{The tuning results of temperature $\tau_1$ in $\mathcal{L}_{PTC}$ on the two datasets.}
    \label{temperature}
\end{figure}

\begin{figure}[t]
    \subfigure[MPTD2]{
       \centering
        \includegraphics[width=0.225\textwidth]{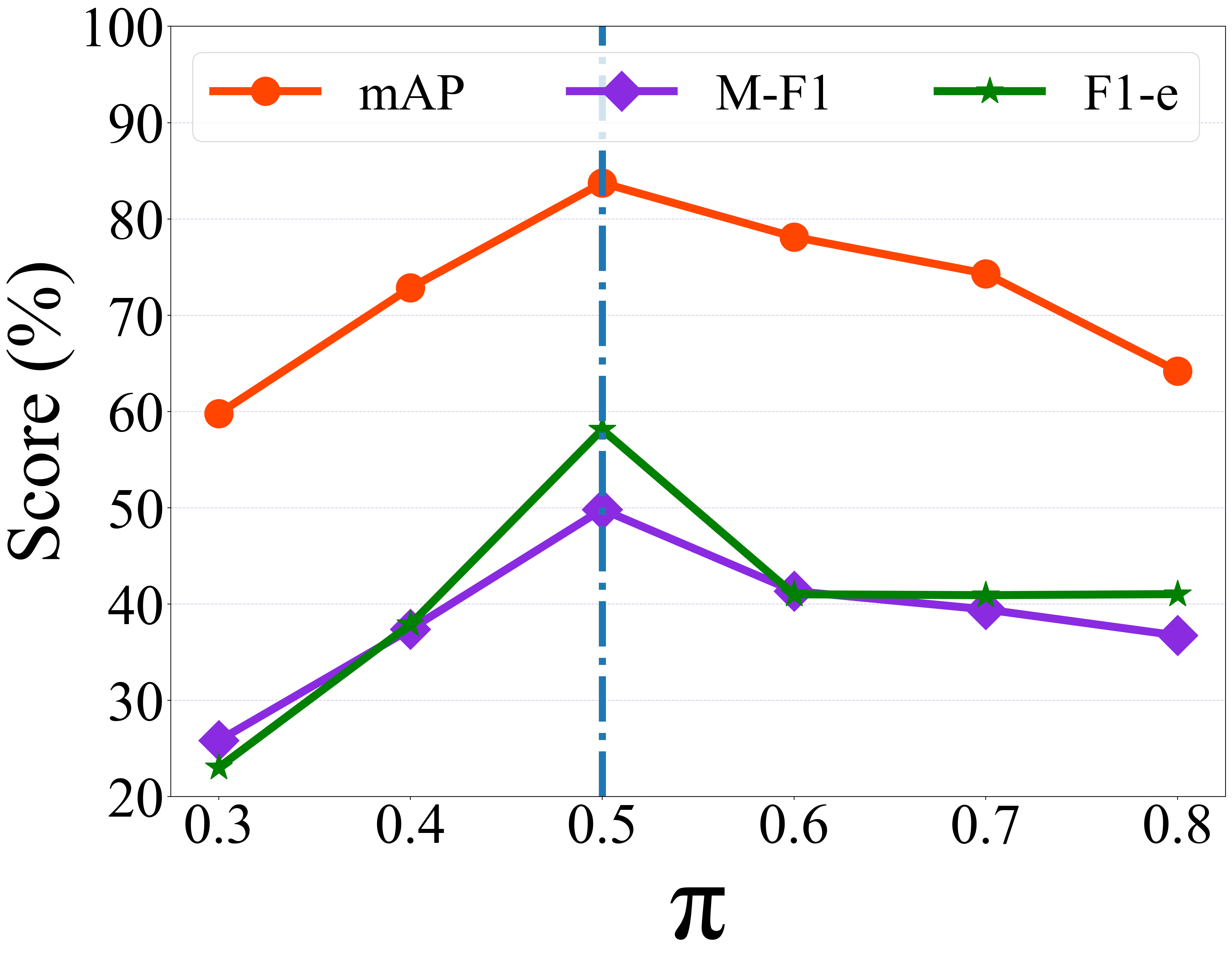}
    }
    \subfigure[MPTD1]{
        \centering
        \includegraphics[width=0.225\textwidth]{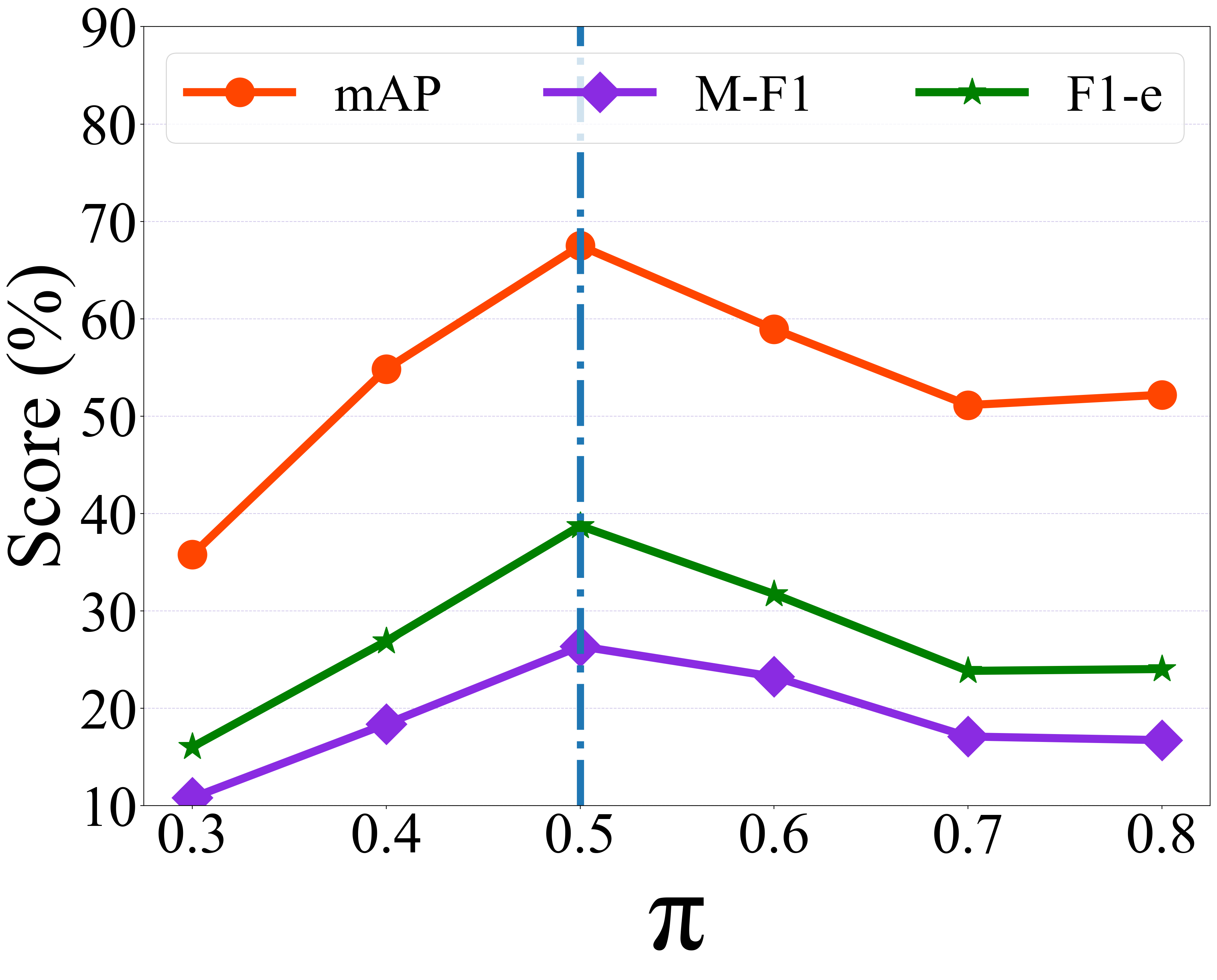}
    }
    \vspace{-0.2cm}
    \caption{The tuning results of prediction threshold $\pi$ on the two datasets.}
    \label{Score}
\end{figure}

We first focus on the impact of the temperature in $s(p,t)$ in Equation \ref{eq:Lptc}, i.e., $\tau_1$. As we mentioned before, the temperature in the scoring function of contrastive learning is used to maintain the balance between the alignment and consistency of contrasted samples. The results depicted in Fig. \ref{temperature} show that M3PT performs better when $\tau_1=0.08\sim 0.12$. It indicates that a suitable temperature should not be too big. Otherwise, it would make the model less focused on discriminating the hard negative samples and thus fail to obtain satisfactory performance. 

We also display the tuning results of the prediction threshold ($\pi$) in Fig. \ref{Score}. It shows that $\pi=0.5$ is the best setting, which just equals the default threshold in generic binary classifications.

\begin{figure}[t]
    \includegraphics[width=3.4in]{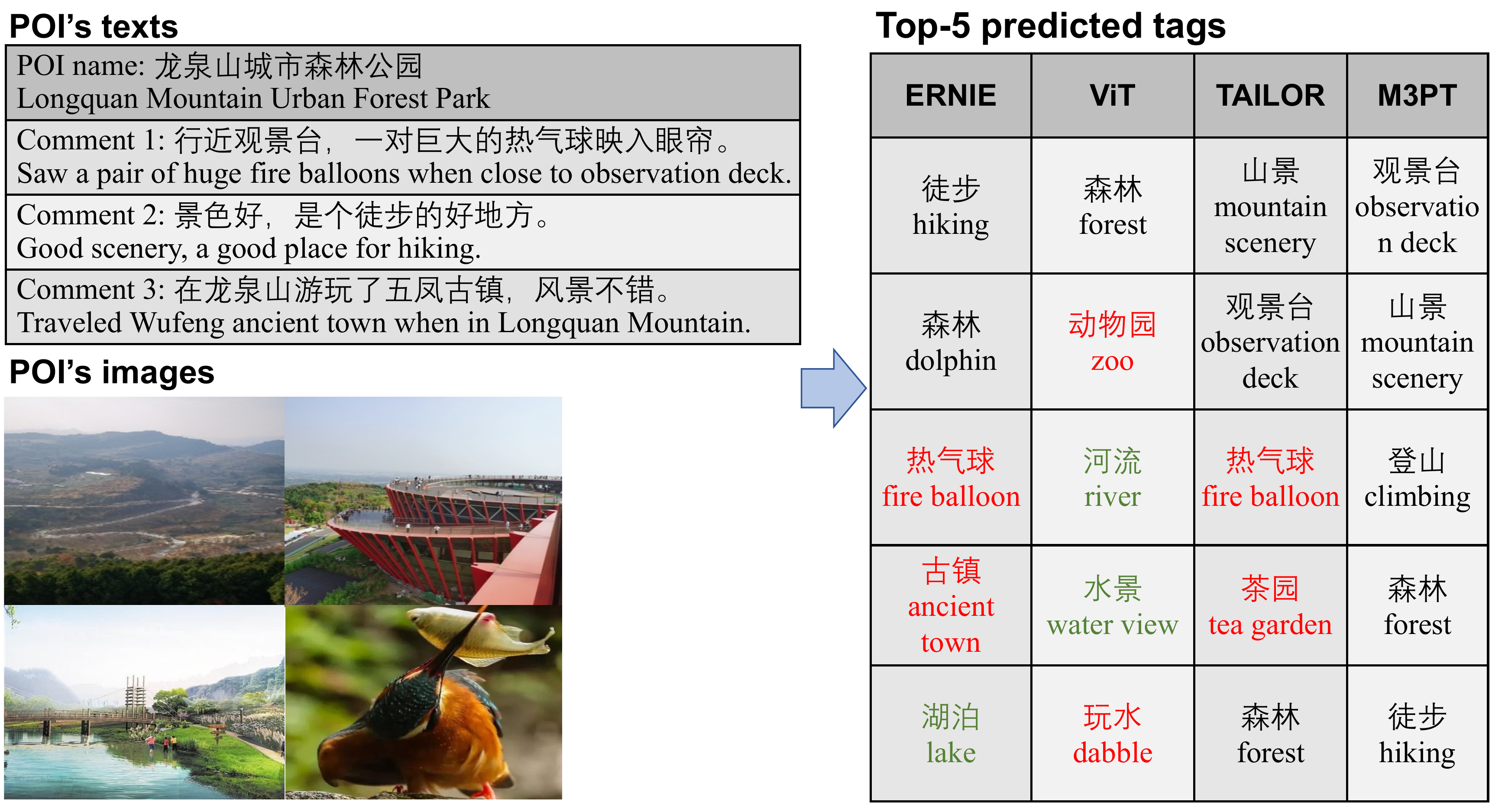}
    \vspace{-0.3cm}
    \caption{The top-5 tags predicted by the models for a POI from MPTD2. The incorrect tags are marked red, the green tags were assessed as correct but not in the POI's gold tag set. And the rest tags are its gold tags (better viewed in color). It shows that M3PT predicts more gold tags.}
    \label{fig:case}
\end{figure}

\begin{CJK}{UTF8}{gbsn}

\subsection{Case Study}

We further display the tagging results of M3PT and the best baseline in each group for a specific POI `龙泉山城市森林公园/Longquan Mountain Urban Forest Park' from MPTD2. The top-5 tags predicted by the compared models are listed in Fig. \ref{fig:case}, where the incorrect tags are marked red, and the green tags were assessed as correct but not in this POI's gold tag set. Moreover, the rest black tags are gold tags. At the same time, we display some texts and images of this POI on the left, which were fed into the models for tag prediction. Obviously, the incorrect tag `热气球/fire balloon' and `古镇/ancient town' were predicted by ERNIR and TAILOR, due to the misleading of Comments 1 and 3. While the incorrect tag `玩水/dabble' and `茶园/tea garden' were predicted by ViT and TAILOR, due to the misleading of some images. In fact, these incorrect tags only have the similar semantics to uni-modal (either textual or visual) data. They can be filtered out through the alignment (matching) of cross-modality. Compared with the baselines, our M3PT achieves the full fusion and precise matching between the textual and visual features, and are less disturbed by the noise of each modal data. Consequently, all of the 5 tags predicted by M3PT are correct.

\end{CJK}

%% file: 060con_correct.tex
Towards the POI tagging in the real-world tour scenario of Ali Fliggy, in this paper, we propose a novel multi-modal model, namely M3PT, which incorporates the textual and visual features of POIs simultaneously to achieve the tagging task. In M3PT, we specially devise a domain-adaptive image encoder (DIE) to generate the image embeddings to better adapt to the requirements of the real-world scenario. In addition, we build the text-image fusion module (TIF) in our model, to achieve the full fusion and precise matching between textual and visual features. We further adopt a contrastive learning strategy to bridge the gap among the corss-modal representations in the model. Our extensive experiments with two datasets that were constructed from the Fliggy platform, not only demonstrate M3PT's advantage, but also justify the rationalities and effectiveness of its important components.

%% file: 070acknowledgements.tex
This work was supported by Alibaba Group through Alibaba Innovative Research Program, Chinese NSF Major Research Plan (No.92270121), Shanghai Science and Technology Innovation Action Plan (No.21511100401), and Shanghai Sailing Program (No. 23YF1409400).